%% file: ijcnlp2017.tex
\documentclass[11pt,letterpaper]{article}
\usepackage{ijcnlp2017}
\usepackage{times}
\usepackage{latexsym}

\ijcnlpfinalcopy

\def\ijcnlppaperid{1090}

\usepackage{xcolor}
\usepackage{listings}
\colorlet{light-gray}{gray!20}
\usepackage{graphicx}
\usepackage{array}
\usepackage{mathtools}
\usepackage{color,soul}

\usepackage[hang, flushmargin]{footmisc}
\title{~\\\vspace{0.7cm}PubMed 200k RCT:\\\vspace{0cm} a Dataset for Sequential Sentence Classification in Medical Abstracts\\\vspace{1cm}}

\author{Franck Dernoncourt\thanks{\hspace{0mm}These authors contributed equally to this work.}\\
	    Adobe Research
	    \\
	    {\tt dernonco@adobe.com}\\
	   \And
	 Ji Young Lee\footnotemark[1]\\
   	MIT\\
   {\tt jjylee@mit.edu}\\}

\date{}

\begin{document}
\maketitle

\vspace*{0.1cm}

\begin{abstract}
We present PubMed 200k RCT\footnote{The dataset is freely available at \url{https://github.com/Franck-Dernoncourt/pubmed-rct}}, a new dataset based on PubMed for sequential sentence classification. The dataset consists of approximately 200,000 abstracts of randomized controlled trials, totaling 2.3~million sentences. Each sentence of each abstract is labeled with their role in the abstract using one of the following classes: background, objective, method, result, or conclusion. The purpose of releasing this dataset is twofold. First, the majority of datasets for sequential short-text classification (i.e., classification of short texts that appear in sequences) are small: we hope that releasing a new large dataset will help develop more accurate algorithms for this task. Second, from an application perspective, 
researchers need better tools to efficiently skim through the literature. Automatically classifying each sentence in an abstract would help researchers read abstracts more efficiently, especially in fields where abstracts may be long, such as the medical field.
\end{abstract}

\input{body_final}

\bibliography{xample}
\bibliographystyle{emnlp_natbib}

\end{document}

%% file: body_final.tex
\vspace{0.0cm}
\section{Introduction}

Short-text classification is an important task in many areas of natural language processing, such as sentiment analysis, question answering, or dialog management. For example, in a  dialog management system, one might want to classify each utterance into dialog acts~\cite{stolcke2000dialogue}.

~\\
\vspace*{-0.5cm}

In the dataset we present in this paper, PubMed 200k RCT, each short text we consider is one sentence. We focus on classifying sentences in medical abstracts, and particularly in randomized controlled trials (RCTs), as they are commonly considered to be the best
source of medical evidence~\cite{moocsysreviews}. Since sentences in an abstract appear in a sequence, we call this task the \textit{sequential sentence classification} task, in order to distinguish it from general text or sentence  classification that does not have any context.

\begin{figure}[!h]
  \centering
  \includegraphics[width=0.45\textwidth]{{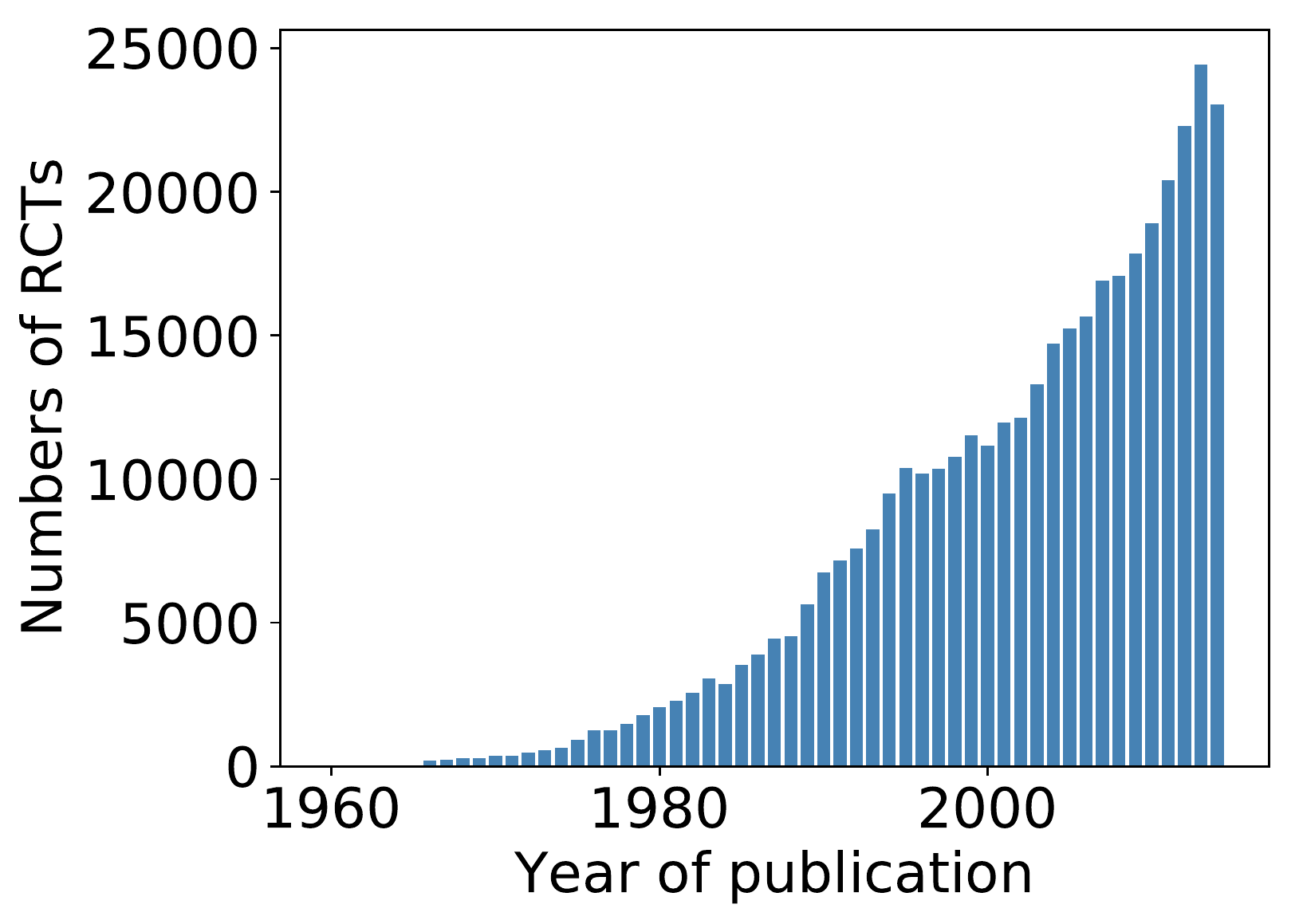}}

  \caption{Number of RCTs present in PubMed published yearly between 1960 and 2014 (inclusive). The first documented controlled trial dates back 1747~\protect\cite{dunn1997james}, but the scientific value of RCTs became widely recognized only by the late 20th   century as the standard method for medical evidence~\protect\cite{meldrum2000brief}.}
  \label{fig:rct-per-year}
\end{figure}

\begin{figure}[!h]
  \centering
  \includegraphics[width=0.45\textwidth]{{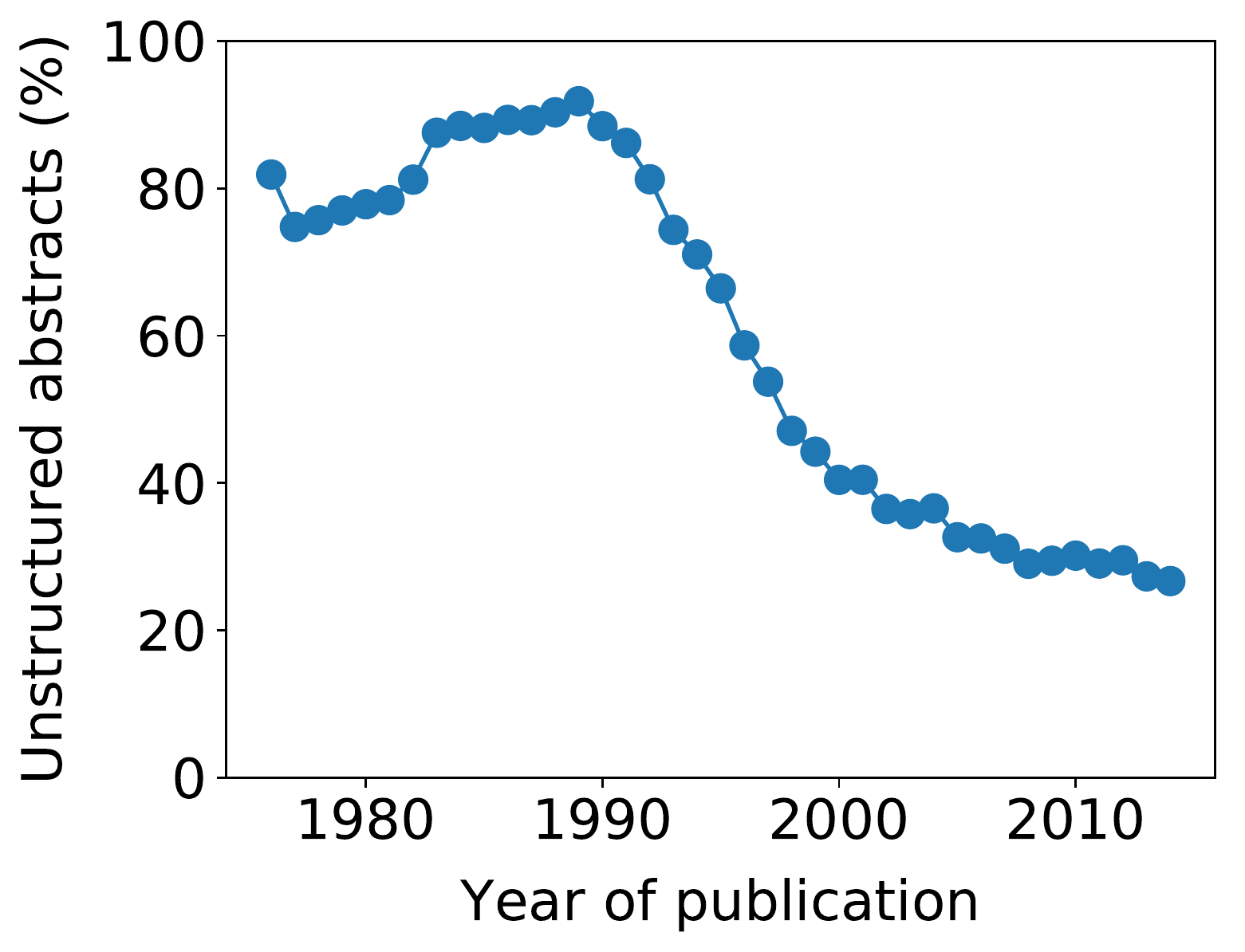}}

  \caption{Evolution of the percentage of RCT abstracts present in PubMed that are unstructured between 1975 and 2014 (inclusive). The years before 1975 were omitted due to the low number of RCTs. Overall, approximately half of the RCT abstracts are unstructured. An RCT abstract is considered as unstructured if and only if at least one of its section is labeled as ``None''.}
  \label{fig:rct-percent-structured}
\end{figure}

The number of RCTs published every year is steadily increasing, as Figure~\ref{fig:rct-per-year} illustrates. Over 1 million 
RCTs have been published so far and around half of them are in PubMed~\cite{mavergames2013}, which makes it challenging for medical investigators to pinpoint the information they are looking for.
When researchers search for previous literature, e.g., to write systematic reviews, they often skim through abstracts in order to quickly check whether the papers match the criteria of interest. This process is easier when abstracts are \emph{structured}, i.e., the text in an abstract is divided into semantic headings such as objective, method, result, and conclusion. However, over half of published RCT abstracts are \emph{unstructured}, as shown in Figure~\ref{fig:rct-percent-structured}, which makes it more difficult to quickly access the information of interest.

\input{tables/highlight-example.tex}

Consequently, classifying each sentence of an abstract to an appropriate heading can significantly reduce time to locate the desired information, as Figure~\ref{tab:highlight-example} illustrates.
 Besides assisting humans, this task may also be useful for a variety of downstream applications such as automatic text summarization, information extraction, and information retrieval.
In addition to the medical applications, we hope that the release of this dataset will help the development of algorithms for sequential sentence classification.

\input{tables/datasets_4pages.tex}

\section{Related Work}

Existing datasets for classifying sentences in medical abstracts are either small, not publicly available, or do not focus on RCTs. Table~\ref{tab:datasets} presents an overview of existing datasets.

The most studied dataset to our knowledge is the NICTA-PIBOSO corpus published by Kim et al.~\shortcite{kim2011automatic}. This dataset was the basis of the ALTA 2012 Shared Task~\cite{amini2012overview}, in which 8~competing research teams participated.

Only the dataset published in~\cite{davis2012detection} is publicly available: 
two datasets can only be obtained via email inquiries, and the other datasets are not accessible (unanswered email requests or negative replies).
The only public dataset is also the smallest one.

\section{Dataset Construction}

\subsection{Abstract Selection}

Our dataset is constructed upon the MEDLINE/PubMed Baseline Database published in 2016
, which we will refer to as PubMed in this paper. PubMed
can be accessed online by anyone, free of charge and without having to go through any registration. It contains 24,358,442 records. A record typically consists of metadata on one article, as well as the article's title and in many cases its abstract.

We use the following information from each PubMed record of an article to build our dataset: the PubMed ID (PMID), the abstract and its structure if available, and the Medical Subject Headings (MeSH) terms. MeSH is the NLM controlled vocabulary thesaurus used for indexing articles for PubMed. 

We select abstracts from PubMed based on the two following criteria: 

\begin{itemize}
\item the abstract must belong to an RCT. We rely on the article's MeSH terms only to select RCTs. Specifically,
only the articles with the MeSH term D016449, which corresponds to an RCT, are included in our dataset. 399,254 abstracts fit this criterion. 
\item the abstract must be structured. In order to qualify as structured, it has to contain between 3 and 9 sections (inclusive), and it should not contain any section labeled as ``None'', ``Unassigned'', or ``'' (empty string). Only 0.5\% of abstracts have  fewer than 3 sections or more than 9 sections: we chose to discard these outliers. The label of each section was originally given by the authors of the articles, typically following the guidelines given by journals: as many labels exist, PubMed maps them into a smaller set of standardized labels: background, objective, methods, results, conclusions, ``None'', ``Unassigned'', or ``'' (empty string). 
\end{itemize}

195,654 abstracts fit these two criteria, i.e.,  belong to RCTs and are  structured.

\subsection{Dataset Split}

\vspace{0.1cm}

The dataset contains 195,654 abstracts and is randomly split into three sets: a validation set containing 2500 abstracts, a test set containing 2500 abstracts,  and a training set containing the remaining 190,654 abstracts.
Since 200k abstracts may be too many for some applications, we also provide a smaller dataset, PubMed 20k RCT, which contains 15000 abstracts for the training set, 2500 abstracts for the validation set, and 2500 abstracts for the test set. The 20k abstracts were chosen from the 200k abstracts by taking the most recently published ones. Table~\ref{tab:datasets-20k-vs-200k} presents the number of abstracts and sentences for both PubMed 20k RCT and PubMed 200k RCT, for each split of the data set. 

\vspace{0.2cm}

\input{tables/datasets_20k-vs-200k.tex}

\subsection{Dataset Format}

\vspace{0.2cm}

The dataset is provided as three text files: one for the training set, one for the validation set, and one for the test set. Each file has the same format: each line corresponds to either a PMID or a sentence with its capitalized label at the beginning. Each token is separated by a space. Listing~\ref{abstract-example} shows an excerpt from these files.

For each abstract, sentence and token boundaries are detected using the Stanford CoreNLP toolkit~\cite{manning-EtAl:2014:P14-5}. We provide two versions of the dataset: one with the original text, and one where digits are replaced by the character @ (at sign).

\vspace{0.3cm}

\begin{lstlisting}[frame=single, backgroundcolor=\color{light-gray}, basicstyle=\footnotesize\ttfamily, language=Java, numbers=none, numberstyle=\tiny\color{black},abovecaptionskip=0.18cm,belowcaptionskip=-0.25cm,caption= {Example of one abstract as formatted in the PubMed 200k RCT dataset set. The PMID of the corresponding article is 9813759; the article can be found that \url{https://www.ncbi.nlm.nih.gov/pubmed/9813759}. },captionpos=b,label={abstract-example}]
###9813759
OBJECTIVE This study evaluated an  [...]
OBJECTIVE It was hypothesized that [...]
METHODS Participants were @ men    [...]
METHODS Psychological functioning  [...]
RESULTS Intervention group subject [...]
RESULTS Compared to the control    [...]
CONCLUSIONS This study has shown   [...]
\end{lstlisting}

\section{Dataset Analysis}

\begin{figure}[!ht]
  \centering
  \includegraphics[width=0.45\textwidth]{{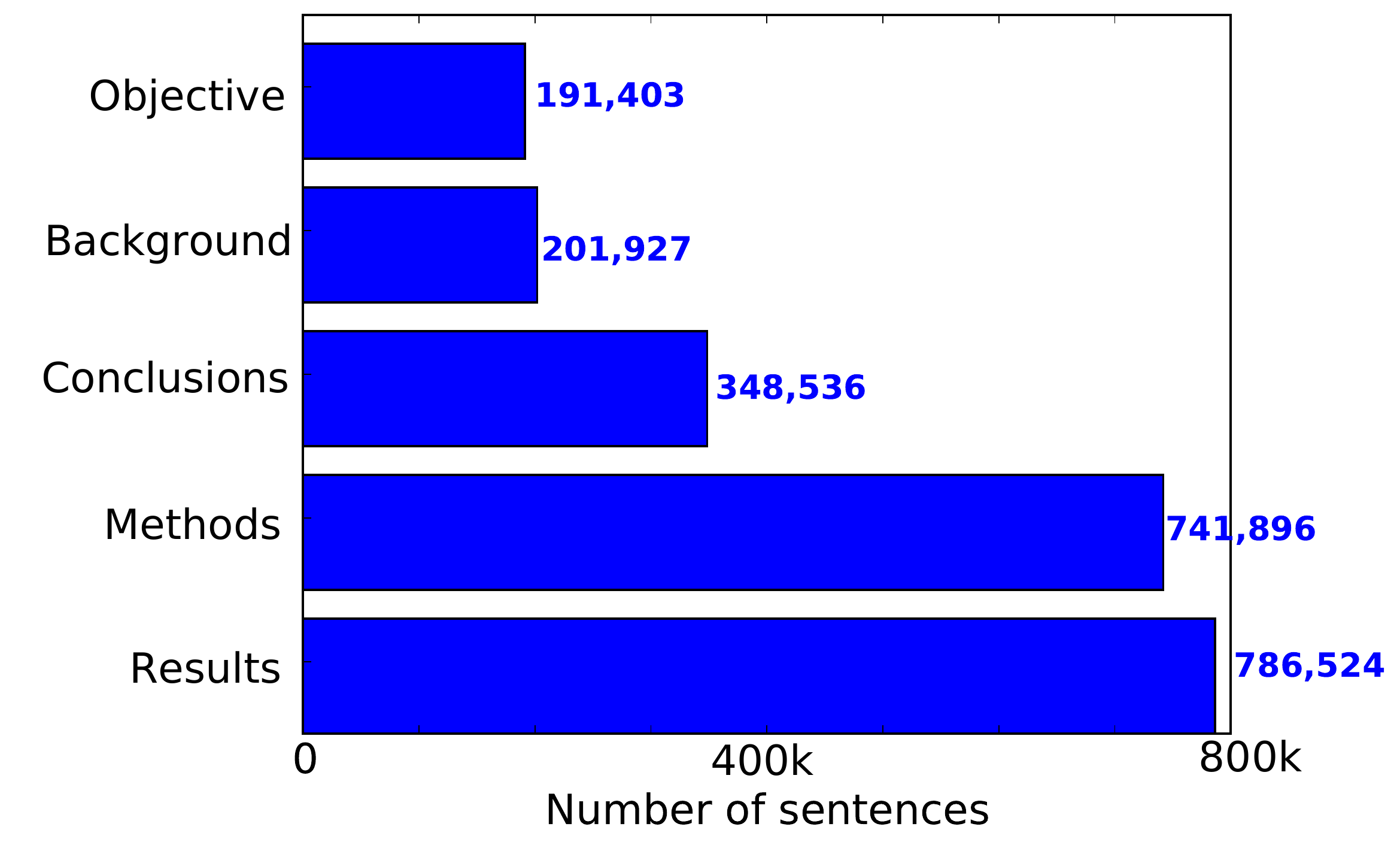}}
  \vspace{-0.4cm}
  \caption{Number of sentences per label}
  \label{fig:label_distribution.pdf}
\end{figure}

\begin{figure}[!ht]
  \centering
  \vspace{-0.3cm}
  \includegraphics[width=0.45\textwidth]{{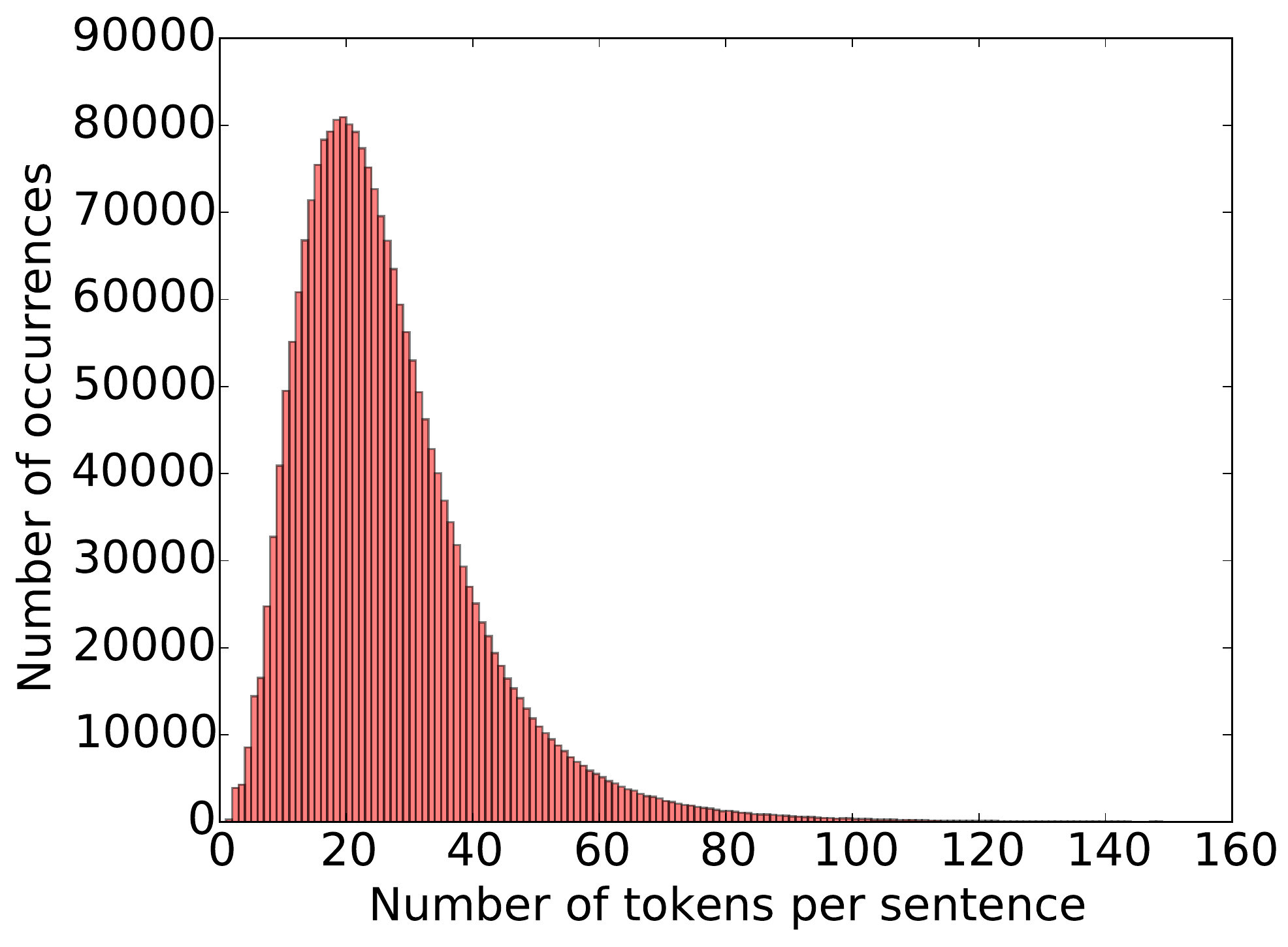}}
  \vspace{-0.4cm}
  \caption{Distribution of the number of tokens the sentence.  Minimum: 1; mean: 26.2; maximum: 338; variance: 227.6; skewness: 2.0; kurtosis: 8.7. }
  \label{fig:text_length_distribution}
\end{figure}

\begin{figure}[!ht]
  \centering
  \vspace{-0.3cm}
  \includegraphics[width=0.45\textwidth]{{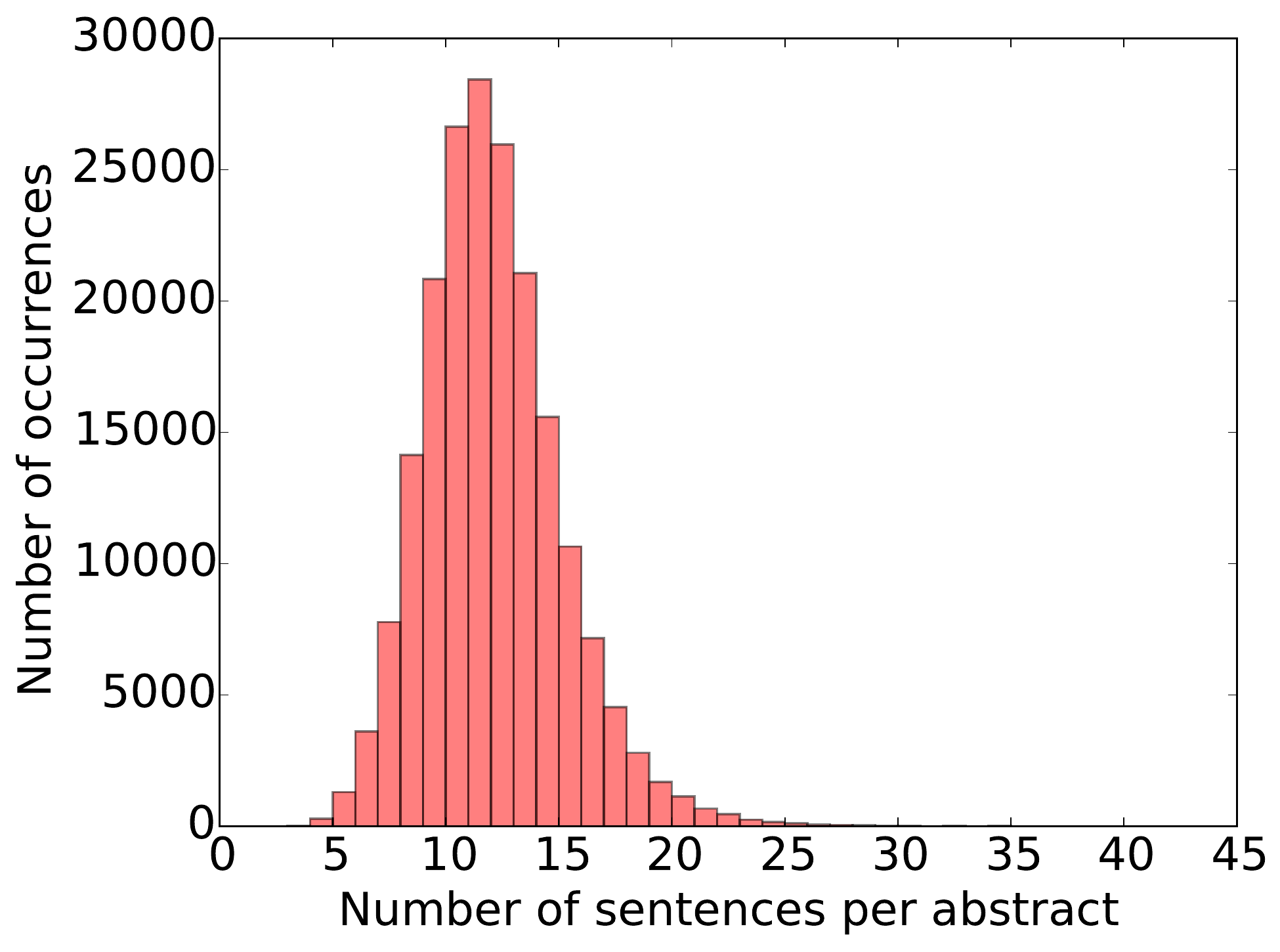}}
  \vspace{-0.4cm}
  \caption{Distribution of the number of sentences per abstract. Minimum: 3; mean: 11.6; maximum: 51; variance: 9.5; skewness: 0.9; kurtosis: 2.6.}
  \label{fig:dialog_length_distribution}
  \vspace{-0.5cm}
\end{figure}

Figure~\ref{fig:label_distribution.pdf} counts the number of sentences per label: the least common label (objective) is approximately four times less frequent than the most common label (results), which indicates that the dataset is not excessively unbalanced.  
Figure~\ref{fig:text_length_distribution} shows the distribution of the number of tokens the sentence. Figure~\ref{fig:dialog_length_distribution} shows the distribution of the number of sentences per abstract. Figures~\ref{fig:label_distribution.pdf},~\ref{fig:text_length_distribution} and~\ref{fig:dialog_length_distribution} are based on PubMed 200k RCT.

\section{Performance Benchmarks}

We report the performance of several systems to characterize our dataset. The first baseline is a classifier based on logistic regression (LR) using n-gram features extracted from the current sentence: it does not use any information from the surrounding sentences. This baseline was implemented with scikit-learn~\cite{pedregosa2011scikit}.

The second baseline (Forward ANN) uses the artificial neural network (ANN) model presented in~\cite{lee2016sequential}: it computes sentence embeddings for each sentence, then classifies the current sentence given a few preceding sentence embeddings as well as the current sentence embedding. 

The third baseline is a conditional random field (CRF) that uses n-grams as features: each output variable of the CRF corresponds to a label for a sentence, and the sequence the CRF considers is the entire abstract.
The CRF baseline therefore uses both preceding and succeeding sentences when classifying the current sentence. CRFs have been shown to give strong performances for sequential sentence classification~\cite{amini2012overview}. This baseline was implemented with CRFsuite~\cite{CRFsuite}. 

The fourth baseline (bi-ANN) is an ANN consisting of three components: a token embedding layer (bi-LSTM), a sentence label prediction layer (bi-LSTM), and a label sequence optimization layer (CRF). The architecture is described in~\cite{dernoncourt2016neural} and 
has been demonstrated to yield state-of-the-art results for sequential sentence classification.

Table~\ref{tab:result-comparisons} compares the four baselines.
As expected, LR performs the worst, followed by the Forward ANN.
The bi-ANN outperforms the CRF, but as the  data set becomes larger the difference of performances diminishes.

Table~\ref{tab:result-details} presents the precision, recall, F1-score and support for each class with the bi-ANN. Accurately classifying the background and objective classes is the most challenging. The confusion matrix in Table~\ref{tab:confusion-matrix} shows that background sentences are often confused with objective sentences, and vice versa.

Table~\ref{tab:result-details-LR} gives more details on the LR baseline, and illustrates the impact of the choice of the n-gram size on the performance.
By the same token, Table~\ref{tab:result-details-CRF} shows the impact of the choice of the window size on the performance of the CRF.

\vspace{3cm}

\input{tables/results-comparisons}

\input{tables/result-details.tex}

\input{tables/confusion-matrix.tex}

\input{tables/result-details-LR.tex}

\input{tables/result-details-CRF.tex}

\section{Conclusion}

In this article we have presented PubMed 200k RCT, a dataset for sequential sentence classification. It is the largest such dataset that we are aware of. We have evaluated the performance of several baselines so that researchers may directly compare their algorithms against them without having to develop their own baselines. We hope that the release of this dataset will accelerate the development of  algorithms  for sequential sentence classification and increase the interest of the text mining community in the study of RCTs.

%% file: tables/highlight-example.tex
\newcommand{\attvis}[2]{\definecolor{att}{rgb}{1, #2, #2} \colorbox{att}{#1}}

\begin{figure}[t]
\small
\centering%
\begin{tabular}{| p{\dimexpr0.47\textwidth-2\tabcolsep-\arrayrulewidth\relax}|   }
\hline
\\
Achilles tendinopathy (AT) is a common and difficult to treat musculoskeletal disorder.  The purpose of this study is to examine whether 1 injection of platelet-rich plasma (PRP) would improve outcomes more effectively than placebo (saline) after 3 months when used to treat AT. 
\hl{A total of 24 male patients with chronic AT (median disease duration, 33 months) were randomized (1:1) to receive either a blinded injection of PRP (n = 12) or saline (n = 12). Patients were informed that they could drop out after 3 months if they were dissatisfied with the treatment.} 
After 3 months, all patients were reassessed (no dropouts). No difference between the PRP and the saline group could be observed with regard to the primary outcome (VISA-A score: mean difference [MD], -1.3; 95\% CI, -17.8 to 15.2; P = .868). Secondary outcomes were pain at rest (MD, 1.6; 95\% CI, -0.5 to 3.7; P = .137), pain while walking (MD, 0.8; 95\% CI, -1.8 to 3.3; P = .544), pain when tendon was squeezed (MD, 0.3; 95\% CI, -0.2 to 0.9; P = .208).
PRP injection did not result in an improved VISA-A score over a 3-month period compared with placebo. 
The conclusions are limited to the 3 months after treatment owing to the large dropout rate. 
\\
\\
\hline

\end{tabular}
\caption{Example of abstract with the method section highlighted. Abstracts in the medical field can be long. This abstract was taken from~\protect\cite{krogh2016ultrasound} and several sentences have been removed for the sake of conciseness. Providing clinical researchers and practitioners a tool that would allow them to highlight the section(s) that
they are interested in would help them explore the literature more efficiently.}
\label{tab:highlight-example}
\end{figure}

%% file: tables/datasets_4pages.tex
\begin{table} [t]
\footnotesize
\centering
\setlength\tabcolsep{4.0pt}
\setlength{\extrarowheight}{3pt}
\setlength{\arraycolsep}{5pt}
\begin{tabular}{|l|c|c|c|c|}
\hline
Dataset & Size 	& Manual & RCT & Available \\
\hline

Hara et al. \protect\shortcite{hara2007extracting}	& 200		&  y & y	& email 	\\

Hirohata et al. \protect\shortcite{hirohata2008}	& 104k		&  n & n	& no	\\

Chung \protect\shortcite{chung2009towards}	& 327	 	& y	&y & no \\

Boudin et al. \protect\shortcite{boudin2010combining}	& 29k  	& 	n& n & no	\\

Kim et al. \protect\shortcite{kim2011automatic}	& 1k		& y	& n& email	\\

Huang et al. \protect\shortcite{huang2011classification}	& 23k &  n 	& n	&  no	\\ 

Robinson \protect\shortcite{robinson2012finding}	&  1k		&n	& y& no	\\ 

Zhao et al. \protect\shortcite{zhao2012exploiting}	& 20k	&  y	& n&  no	\\

Davis et al. \protect\shortcite{davis2012detection}& 194	 &  n & y	&  public
\\ 

Huang et al. \protect\shortcite{huang2013pico}	& 20k	&  n & y	&  	no\\ 

PubMed 200k RCT	& 196k 		&  n & y	&  	no\\

\hline
\end{tabular}
\caption{Overview of existing datasets for sentence classification in medical abstracts. The size is expressed in terms of number of abstracts. 
} \label{tab:datasets}
\end{table}

%% file: tables/datasets_20k-vs-200k.tex
\begin{table} [t]
\footnotesize
\centering
\setlength\tabcolsep{3.0pt}
\setlength{\extrarowheight}{3pt}
\setlength{\arraycolsep}{0pt}
\begin{tabular}{|l|c|c|c|c|}
\hline
\textbf{Dataset} &  \textbf{$|V|$} 	& Train & Validation & Test \\
\hline
\text{PubMed 20k}	& 	68k	& 15k (180k)	& 2.5k (30k) 	& 2.5k	(30k)	\\ %
\text{PubMed 200k}			&	331k &	190k (2.2M)		& 2.5k (29k) 	& 200 (29k)	 \\ 
\hline
\end{tabular}
\caption{Dataset overview.   
$|V|$ denotes the vocabulary size. For the train, validation and test sets, we indicate the number of abstracts followed by the number of sentences in parentheses.
} \label{tab:datasets-20k-vs-200k}
\end{table}

%% file: tables/results-comparisons.tex
\begin{table} [!t]
\footnotesize
\centering
\setlength{\extrarowheight}{3pt}
\setlength{\arraycolsep}{5pt}
\begin{tabular}{|l|c|c|c|c|c|c|}
\hline
\textbf{Model} & PubMed 20k	& PubMed 200k \\
 \hline

LR			& 83.1		&	85.9	 \\
Forward ANN		& 86.1			&	88.4	 \\
CRF		& 89.5			&	91.5	 \\
bi-ANN		& \textbf{90.0}			&	\textbf{91.6}	 \\
\hline
\end{tabular}
\caption{F1-scores on the test set of several baselines. The presented results for the ANN-based models are the F1-scores on the test set of the run with the highest F1-score on the validation set.\vspace{-0.1cm}
} \label{tab:result-comparisons}
\end{table}

%% file: tables/result-details.tex
 \begin{table}[!h]

\footnotesize
\centering
\setlength\tabcolsep{6.0pt}
\setlength{\extrarowheight}{3pt}
\setlength{\arraycolsep}{5pt}
\begin{tabular}{|l|cccc|}
\hline 
\cline{2-5} 
 & Precision & Recall & F1-score & Support \tabularnewline
\hline 
Background	& 70.7	& 81.1	& 75.6 	& 2663  \tabularnewline
Conclusions	& 94.6	& 93.7	& 94.2	& 4426 \tabularnewline
Methods		& 95.5	& 96.5	& 96.0	& 9751 \tabularnewline
Objective	& 77.1	& 65.3	& 70.7	& 2377 \tabularnewline
Results	 	& 95.6 	& 94.8	& 95.2	& 10276 \tabularnewline
\hline
Total	 	& 91.7	& 91.6	&  91.6	& 29493\tabularnewline
\hline 
\end{tabular}
\caption{Results for each class obtained by the bi-ANN model on the PubMed 200k RCT test set. The total support is 29493, i.e. the number of sentences in the test set. 
} 
\vspace{0.2cm}
\label{tab:result-details}
\end{table}

%% file: tables/confusion-matrix.tex
\begin{table} [!h]
\footnotesize
\centering
\setlength\tabcolsep{4.0pt}
\setlength{\extrarowheight}{3pt}
\setlength{\arraycolsep}{5pt}
\begin{tabular}{|l|c|c|c|c|c|}
\hline
  & Backg. 	& Concl. 	& Methods & Obj. & Res. \\
\hline
Background		& 	2760	& 12	 & 	62		& 424 	& 5	\\ %
Conclusions		& 	41	& 4149	 &	9		& 0 	& 227	 \\
Methods			& 	82	& 17	 &	9409		& 31 	& 212	 \\
Objective		& 	757 	& 0	 &	69		&  1551	& 0	 \\
Results			& 14	& 208	 &	303		&  5	& 9746		 \\
\hline
\end{tabular}
\caption{Confusion matrix on the PubMed 200k RCT test set obtained with the bi-ANN model. Rows correspond to actual labels, and columns correspond to predicted labels. For example, 62 background sentences were predicted as method.} \label{tab:confusion-matrix}
\end{table}

%% file: tables/result-details-LR.tex
 \begin{table}[!t]

\footnotesize
\centering
\setlength\tabcolsep{6.0pt}
\setlength{\extrarowheight}{3pt}
\setlength{\arraycolsep}{5pt}
\begin{tabular}{|c|cccc|}
\hline 
\cline{2-5} 
N-gram size & Precision & Recall & F1-score & Runtime \tabularnewline
\hline 
1	& 82.3	& 82.7	& 82.4 	& 4406  \tabularnewline
2	& 85.1	& 85.4	& 85.2 	& 13237  \tabularnewline
3	& 85.5	& 85.8	& 85.6 	& 20618  \tabularnewline
4	& 85.7	& 86.0	& 85.8 	& 25553 \tabularnewline
5	& \textbf{85.8}	& \textbf{86.1}	& \textbf{85.9} 	& 35006 \tabularnewline
\hline 
\end{tabular}
\caption{Results obtained on the PubMed 200k RCT test set by the LR model with different size of n-grams as features. The n-gram size indicates the size of the largest n-grams: For example, if the n-gram size is 3, it means unigrams, bigrams and trigrams are extracted as features. The maximum n-gram size in our experiments is 5 due to RAM limitation. The runtime is expressed in seconds and comprises both training and testing times.} 
\label{tab:result-details-LR}
\end{table}

%% file: tables/result-details-CRF.tex
 \begin{table}[!t]

\footnotesize
\centering
\setlength\tabcolsep{6.0pt}
\setlength{\extrarowheight}{3pt}
\setlength{\arraycolsep}{5pt}
\begin{tabular}{|c|cccc|}
\hline 
\cline{2-5} 
Window size & Precision & Recall & F1-score & Runtime \tabularnewline
\hline 
1	& 90.6	& 90.6	& 90.6 	& 1565 	\tabularnewline
2	& 91.0	& 91.0	& 91.0 	& 2490    \tabularnewline
3	& 91.1	& 91.1	& 91.1 	& 3908    \tabularnewline
4	& \textbf{91.5}	& \textbf{91.5}	& \textbf{91.5} 	& 4867    \tabularnewline
5	& 90.9	& 91.0	& 90.9 	& 6424    \tabularnewline
6	& 91.4	& 91.4	& 91.4 	& 7649    \tabularnewline
7	& 91.3	& 91.3	& 91.3 	& 7929    \tabularnewline
8	& 90.9	& 90.9	& 90.9 	& 7644    \tabularnewline
9	& 91.2	& 91.3	& 91.2 	& 7891    \tabularnewline
\hline 
\end{tabular}
\caption{Results obtained on the PubMed 200k RCT test set by the CRF model with different window sizes. A window of size $k$ means that for each token, features are extracted from the current token, the $k$ preceding tokens as well as the $k$ succeeding tokens.  The runtime is expressed in seconds and comprises both training and testing times.} 
\label{tab:result-details-CRF}
\end{table}